\documentclass[11pt]{article}

\usepackage[a4paper,margin=1in]{geometry}
\usepackage{times}
\usepackage{microtype}
\usepackage{booktabs}
\usepackage{multirow}
\usepackage{enumitem}
\usepackage{amsmath,amssymb}
\usepackage{hyperref}
\usepackage{graphicx}
\usepackage{url}
\usepackage{algorithm}
\usepackage{algorithmic}
\usepackage{listings}
\usepackage{xcolor}
\usepackage{mathtools}
\usepackage{caption}
\usepackage{subcaption}
\usepackage{csquotes}
\usepackage{footmisc}
\usepackage{pifont}
\usepackage{url}

\usepackage{xcolor}
\definecolor{darkgreen}{rgb}{0.0, 0.5, 0.0}
\definecolor{darkred}{rgb}{0.8, 0.0, 0.0}
\usepackage{makecell}

\definecolor{lightgray}{gray}{0.95}
\usepackage{tabularx}
\usepackage{booktabs}

\usepackage{tikz}
\usetikzlibrary{shapes.geometric, arrows.meta, positioning, fit, calc}

\title{"Don't Teach Minerva": Guiding LLMs Through Complex Syntax for Faithful Latin Translation with RAG}

\author{
  {Sergio Torres Aguilar (sertor01@ucm.es)}}

\date{}

\begin{document}
\maketitle

\begin{abstract}
Translating a morphology-rich, low-resource language like Latin poses significant challenges. This paper introduces a reproducible draft-based refinement pipeline that elevates open-source Large Language Models (LLMs) to a performance level statistically comparable to top-tier proprietary systems. Our method first uses a fine-tuned NLLB-1.3B model to generate a high-quality, structurally faithful draft. A zero-shot LLM (Llama-3.3 or Qwen3) then polishes this draft, a process that can be further enhanced by augmenting the context with retrieved out-context examples (RAG). We demonstrate the robustness of this approach on two distinct benchmarks: a standard in-domain test set (Rosenthal, 2023) and a new, challenging out-of-domain (OOD) set of 12th-century Latin letters (2025). Our central finding is that this open-source RAG system achieves performance statistically comparable to the GPT-5 baseline, without any task-specific LLM fine-tuning. We release the pipeline, the Chartres OOD set, and evaluation scripts and models to facilitate replicability and further research.

\end{abstract}

\section{Introduction}

Machine translation (MT) of Latin faces a unique confluence of challenges. The language itself is inherently difficult, with its rich morphology, complex case-markings, and flexible word order. This linguistic complexity is then magnified by a vast diachronic landscape: a textual tradition spanning two millennia has produced an immense diversity of styles, from classical prose to medieval law and Humanistic neolatin, that no single model can easily master. Yet, the high-quality parallel corpora needed for training are both scarce and stylistically narrow, often relying on archaic English references that penalize modern surface-level metrics like BLEU~\cite{papineni2002bleu}, while the scholarly practice of permitting multiple legitimate interpretations makes any single reference a sparse and potentially biased target. While massive proprietary language models have set a new standard for zero-shot translation, they present their own challenges regarding reproducibility, cost, and control for specialized scholarly applications that demand philological precision within budget limitations.

This paper addresses a key question: can a fully open-source and reproducible pipeline achieve performance comparable to these state-of-the-art proprietary systems? We argue that it can, by strategically combining the strengths of specialized models and general-purpose reasoners. We propose a two-stage pipeline: a fine-tuned NMT model provides a domain-specialized draft, which a zero-shot Large Language Model (LLM) then refines, guided by relevant in-context examples retrieved via semantic search. Our core hypothesis is that this method of providing targeted, inference-time guidance allows open-source LLMs to be on par with top-tier proprietary models, particularly on challenging, out-of-domain texts.

\paragraph{Contributions.}
\begin{itemize}[leftmargin=1.2em]
    \item We introduce a transparent, two-stage refinement pipeline for Latin-to-English translation. The system refines drafts from a specialized NLLB model \cite{nlbb2022} with a zero-shot LLM (Llama-3.3 \cite{dubey2024llama} or Qwen3 \cite{yang2025qwen3}), achieving state-of-the-art results.
    \item We introduce and evaluate performance on a new, challenging out-of-domain (OOD) test of 12th-century Latin letters, based on contemporary scholarly translations \cite{Yves_Chartres} in a diplomatic style.
    \item We demonstrate that our open-source Llama-3.3-70B+RAG system reaches the GPT-5 baseline on the ID and OOD benchmarks without specific fine-tuning, validating the viability of guided open models for specialized MT tasks.
    \item A detailed error taxonomy and qualitative analysis highlighting philological adequacy (e.g., handling of dative/ablative roles, negation, tense) alongside traditional metrics. 
\end{itemize}

\section{Related Work}\label{sec:related}
\textbf{Latin NLP and resources.}
While Classical Latin benefits from linguistic resources like treebanks and lexica (e.g., PROIEL, Perseus/CLTK), Latin–English parallel data remain scarce. Rosenthal ~\cite{rosenthal2023latin} assembled ~100k pairs spanning Vulgate and Loeb classical texts; OPUS Bible adds ~63k pairs \cite{opusbible}, but they are largely drawn from 19th-century English references. This creates a domain and style mismatch for much of the Latin corpus, a challenge our new medieval OOD test set is designed to address. The diachronic depth of Latin and editorial variance complicate alignment and evaluation using surface-oriented MT metrics and models that prefer English-like word order, motivating our use of semantic metrics and detailed qualitative analysis.

\textbf{Neural MT for Historical Languages.}
Traditional rule-based pipelines and early statistical systems offered coverage but struggled with complex syntax ~\cite{white2015blitz}. Domain-adaptive pre-training (DAPT)~\cite{iyer2024quality} can help low-resource MT~\cite{kocmi2021one}, but overfitting to reference style remains a risk. Many established NMT pipelines rely on seq2seq architectures like NLLB and mBART. More recently, proprietary LLMs have redefined the state-of-the-art, with models like GPT-4 achieving significant gains in zero-shot translation over previous NMT systems~\cite{volk-etal-2024-llm}. However, the potential of applying advanced prompting techniques to open-source LLMs for Latin has remained largely unexplored. Our work fills this gap, demonstrating that a zero-shot RAG approach could be competitive with a strong proprietary ML baseline.

\textbf{Retrieval augmentation and translation memory}.
RAG and translation-memory-style prompting improve factuality and terminology by injecting context at inference. For MT, retrieval has been shown to stabilize lexical choice and local syntax, but Latin poses a particularly clause-centric challenge: ablative absolutes, case-licensed predicates, and relative-clause anchoring. We operationalize retrieval at this level by pairing a draft NMT (for coverage) with a zero-shot refiner guided by exemplars of semantically similar source texts and their corresponding NLLB drafts.

\textbf{Latin MT benchmarks and recent systems.}
For Latin-English, Rosenthal~\cite{rosenthal2023latin} and Fischer et al.~\cite{fischer2022machine} established strong NMT baselines with BLEU scores in the low 20s in 2023. Proprietary models soon pushed this boundary, with GPT-4 reaching a BLEU of 34.5~\cite{volk-etal-2024-llm} on 16th Latin texts. More recently, Littera (2025)~\cite{rosu-2025-litera} reported exceptionally high scores using a complex, multi-call proprietary pipeline (12 LLM calls), though its reliance on a tiny, custom test set and massive computational overhead makes comparison difficult. Our work positions itself differently: we aim for an efficient and fully open-source system that proves its robustness against a strong, single-call proprietary baseline (GPT-5 \cite{gpt-5}) across multiple, distinct benchmarks, including a new OOD set.


\section{Methodology}\label{sec:models}

Our approach is centered on a two-stage, retrieval-augmented pipeline designed to leverage the complementary strengths of a specialized NMT model and a general-purpose LLM. This section details the models used, the pipeline architecture, and the training protocol for our specialized components.

\subsection{Model Architecture and Baselines}
We evaluate four model families to cover a range of architectures, sizes, and access paradigms, as summarized in Table~\ref{tab:model-props}. Our strategy involves three distinct roles:
\begin{itemize}[leftmargin=1.2em]
\item \textbf{A specialized drafter:} A fine-tuned NLLB-200-1.3B, a traditional encoder-decoder NMT model, chosen for its strong multilingual foundation and efficiency.
\item \textbf{Open-source refiners:} Decoder-only LLMs of varying scales (Llama-3.3-70B, Qwen3-14B/30B) used in a zero-shot capacity to preserve their general reasoning capabilities.
\item \textbf{A proprietary benchmark:} A single-call API to GPT-5 to contextualize our results against the current state-of-the-art.
\end{itemize}

\begin{table}[H]
\centering
\small
\setlength{\tabcolsep}{5pt}
\begin{tabular}{lccccccc}
\toprule
\textbf{Model} & \textbf{Params} & \textbf{Arch.} & \textbf{Open wts.} & \textbf{Multilingual} & \textbf{Type}  & \textbf{Release} \\
\midrule
NLLB-200-1.3B & 1.3B & Enc--Dec & \checkmark & 200 & seq2seq &2023-05 \\
\midrule
Llama-3.3-70B & 70B & Dec-only & \checkmark\!* & Strong (8 main) & Instruct & 2024-12 \\
Qwen3-14B & 14B & Dec-only & \checkmark & Strong (+100) & Thinking & 2025-05 \\
Qwen3-30B & 30B$^\dagger$ & Dec-only & \checkmark & Strong (+100) & Instruct & 2025-05 \\
\midrule
GPT-5 (\footnotesize{API}) & $>$2T? & Undisclosed & \ding{55} & Strong (+100) & Thinking & 2025-08 \\
\bottomrule
\end{tabular}

\caption{Model families compared and key properties.
\!* gated download under Meta’s community terms.
$^\dagger$ Mixture-of-Experts with approximately 3 billion active parameters per forward pass.} 
\label{tab:model-props}
\end{table}

\subsection{Two-Stage Retrieval-Augmented Refinement}

\textbf{Stage 1: Draft Generation.} Our fine-tuned NLLB-1.3B model first produces an initial, structurally faithful draft of the source Latin text.

\textbf{Stage 2: Retrieval and Augmented Refinement.} The refiner LLM receives the original Latin and this NLLB draft ($k=1$). For augmentation, we retrieve the top 5 most similar Latin neighbors from our 50M-token corpus and generate their drafts on-the-fly with the same NLLB model ($k>1$). The final prompt instructs the LLM to polish the main draft using these five (neighbor, draft) pairs for guidance.

\textbf{Rationale.}
This two-stage process is deliberate. The NLLB draft acts as an \emph{ad verbum} anchor, effectively conditioning the LLM refiner, which better handles instructions and refinement, to a controlled post-editing task. The final prompt, shown below, instructs the LLM to refine the provided semantically enriched draft into a single, faithful English translation.

\textbf{Prompt shape (refiner).}
The top $k=>1$ Neighbors are provided as analogical exemplars (Latin + NLLB draft). The model outputs a \emph{single} English line :
\begin{lstlisting}[frame=tb, backgroundcolor=\color{white}, rulecolor=\color{black!20}]
System: You are an expert classicist translator. Produce ONE faithful,
English translation. Preserve case roles and polarity. No extra text.

User:
- Revise the Draft Translation to be a more accurate and fluent version of the Latin source text.

Latin text: <latin>

NMT draft (NLLB): <draft>

Use the Analogous examples for guidance:
[EX1] LATIN: <neighbor_latin>
[EX1] DRAFT: <neighbor_draft>
... (k=5)

Final translation:
\end{lstlisting}

\begin{figure*}[h!]
\centering
\scalebox{0.21}{
\includegraphics{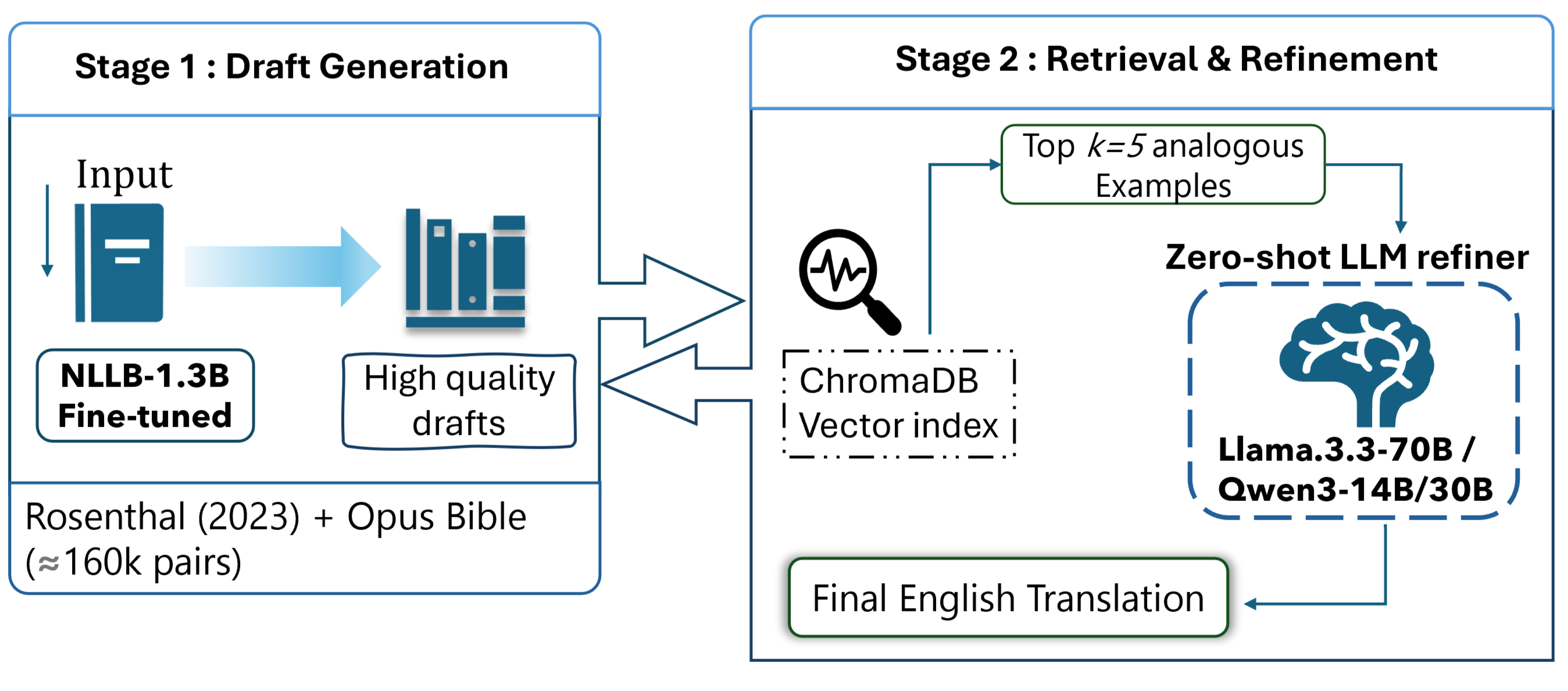}}
 \caption{Overview of our two-stage RAG pipeline. A specialized NLLB model first generates a high-quality draft. This draft is then refined by a zero-shot LLM, which is augmented with in-context examples retrieved from a non-parallel corpus via semantic search.}
\label{fig:pipeline}
\end{figure*}

\subsection{Training Protocol}
\textbf{NLLB Domain Adaptation.} Our only trained component is the NLLB-1.3B drafter, which undergoes two phases. First, we perform domain-adaptive pre-training (DAPT) using LoRA on our 50M-token raw Latin corpus. Second, the adapted model is fully fine-tuned on our ~160k-pair parallel corpus (Rosenthal + OPUS Bible) to specialize it for Latin-to-English translation.

\textbf{Zero-Shot LLM Refiners.} The Llama and Qwen models are used without any task-specific fine-tuning and no gradient updates. This is a deliberate choice to preserve their generalization capabilities.

\subsubsection{Hyperparameters}
\begin{itemize}[leftmargin=1.2em]
\item \textbf{DAPT LoRA:} rank 16, $\alpha{=}32$, attention+MLP, LR 
$2{\times}10^{-4}$ cosine, bf16, seq len 1024.
\item \textbf{NLLB FT:} LR $5{\times}10^{-5}$, 3 epochs, completion-only loss, standard seq2seq template.
\item \textbf{RAG:} $k{=}5$ neighbors, embedding \texttt{BAAI/bge-m3} \cite{bge_m3}, Gen. Temp. $0.0$ (ID), $0.0/0.5$ (OOD).
\item \textbf{Hardware}: 2X GPU (RTX A6000 48GB); batches: 128 (NLLB), 16 (Qwen3-14B), 4 (LLama3.3)
\item \textbf{Libraries}: Transformers 4.55.1; Peft 0.17.1. 4-bit bitsandbytes. 
\item \textbf{Sequences length:} input (1100-1300 tokens for $k=1+5$), output (256)
\end{itemize}

\subsection{Inference Algorithm}
We implement retrieval-augmented refinement as a light-weight wrapper around the refiner API:

\begin{enumerate}[leftmargin=1.2em,itemsep=2pt,topsep=2pt]
    \item \textbf{Draft}: $d \leftarrow \text{NLLB}(x)$ where $x$ is the Latin source.
    \item \textbf{Retrieve Neighbors}: $N \leftarrow \text{TopK}_{\mathcal{D}_{\text{retrieval}}}(\text{Embed}(x), k{=}5)$ with lemma Jaccard $\geq 0.3$ 
    \item \textbf{Assemble Context}: $C \leftarrow \{(\text{LATIN}=n_i,~\text{DRAFT}=\text{NLLB}(n_i))\}_{i=1}^{k}$ for $n_i \in N$.
    \item \textbf{Refine}: $\hat{y} \leftarrow \text{LLM}_{\theta}(\text{SystemPrompt}, x, d, C)$.
\end{enumerate}

\paragraph{Cost comparison.}(10-2025)
\begin{itemize}
    \item From our usage logs, GPT-5 ($k=0$) costs about \textbf{\$1.16} per 100 sentences (23k input, 191k output tokens at \$1.25M/\$10M rates), or \textbf{\$0.58} with batching (asynchronous). 
    
    \item Local: Llama-3.3-70B+RAG system on a single A6000 GPU runs at \textbf{\$0.15--0.20} for the same workload ($\approx$18 minutes), assuming GPU amortization at \$0.50/h plus \$0.10/kWh, excluding labor. 

    \item Cloud: Assuming similar hardware (A6000 + 10vCPU) at \$0.50--0.79 / hr : \textbf{\$0.15--0.24}

\end{itemize}

Thus local inference is roughly \textbf{3--6$\times$ cheaper} at observed output lengths, though costs remain sensitive to output size and batching efficiency.


\section{Experimental Setup}
This section details the datasets, metrics, and settings used to evaluate our pipeline.

\subsection{Corpora}\label{sec:data}
\begin{itemize}[leftmargin=1.2em]
\item \textbf{Parallel Corpus (~160k pairs):} A combination of the Rosenthal classical/biblical corpus \cite{rosenthal2023latin} and the OPUS bible-uedin corpus ~\cite{opusbible}. This pool is used exclusively to fine-tune the NLLB drafter model.

\item \textbf{Domain-Adaptation Corpus (50M tokens):} A collection of non-parallel Latin texts from classical, medieval, and early modern collections. This corpus serves two critical functions: it is used for the initial DAPT phase of the NLLB model, and it serves as the retrieval index for our RAG pipeline, providing a vast pool of semantically rich contexts. (OOD test set is excluded to prevent data leakage).

\item \textbf{In-Domain Test Set (Rosenthal):} The standard test split from Rosenthal (2023), stylistically aligned with the parallel training data.

\item \textbf{Out-of-Domain Test Set (Yves de Chartres):} Our new 110-sentence test set (3500 tokens) from the letters of the bishop Yves de Chartres (1090-1115), using contemporary scholarly translations as references. \cite{Yves_Chartres}.
\end{itemize}

\subsection{Evaluation Metrics and Settings}

We report standard lexical metrics (SacreBLEU~\cite{post2018call}, chrF++~\cite{popovic2015chrf}) and semantic metrics (BERTScore~\cite{zhangbertscore}, COMET-22~\cite{rei2022comet}), both using the multilingual XLM-RoBERTa model). Semantic metrics are better for Latin, as they capture meaning equivalence even when surface forms diverge from the archaic references. Besides, COMET is fine-tuned on human ratings of machine translations. For inference, we used greedy decoding (temperature 0.0), but also explored temperature 0.5 on the OOD set to assess its impact on fluency. The GPT-5 baseline was prompted with a simple, direct instruction ("Translate the following Latin text to English:") and greedy decoding (temp=0.0, top\_p=1.0). We used the GPT-5-large model version available in October 2025.



\section{Experiments}\label{sec:experiments}

\subsection{Results on the In-Domain Test Set}

\begin{table}[h!]
\centering
\small
\begin{tabular}{lcccccc}
\toprule
\textbf{System (Rosenthal Test set)} & \textbf{BLEU}$\uparrow$ & \textbf{chrF++}$\uparrow$ & \textbf{\small{METEOR}}$\uparrow$ & \textbf{BERTScore}$\uparrow$ & \textbf{COMET}$\uparrow$ & \textbf{Speed}$\downarrow$ \\
\midrule
\textit{Baselines}  \footnotesize{(GPU)} & & & & & \\
NLLB-1.3B (fine-tuned) & 25.61 & 48.17 & 0.539 & 0.920 & 0.701 & 1.1 \\
Llama-3.3-70B (zero-shot) & 20.50 & 45.73 & 0.493 & 0.917 & 0.688 & 15.2\\
Qwen3-30B (zero-shot) & 19.12 & 43.50 & 0.473 & 0.911 & 0.673 & 8.2\\
Qwen3-14B (zero-shot) & 19.35 & 44.01 & 0.479 & 0.912 & 0.674 & 5.4 \\
\midrule
\textit{Proprietary Models} \footnotesize{(API)} & & & & & \\
GPT-5-mini & 22.74 & 47.35 & 0.524 & 0.919 & 0.722 & 11.5\\
GPT-5 & 25.23 & \textbf{49.87} & \textbf{0.548} & 0.922 & 0.730 & 16.9\\
\midrule
\textit{Open RAG Systems} \footnotesize{(GPU)} & & & & & \\
Llama-3.3-70B + RAG & \textbf{26.55} & 49.86 & \textbf{0.548} & \textbf{0.925} & \textbf{0.743} & 18.3 \\
Qwen3-30B + RAG & 24.17 & 48.42 & 0.523 & 0.923 & 0.738 & 11.6 \\
Qwen3-14B + RAG & 24.97 & 48.79 & 0.532 & 0.923 & 0.728 & 8.6 \\
\bottomrule
\end{tabular}
\caption{Results on the in-domain Rosenthal test set. Speed is expressed as the time in minutes required to process 100 items at batch 1. We measure end-to-end latency per sentence, including draft generation, embedding, ANN retrieval and filtering, neighbor draft generation and LLM decoding.}
\label{tab:rosenthal-results}
\end{table}

In-domain test results are summarized in Table~\ref{tab:rosenthal-results}. As expected, zero-shot LLMs are outperformed by the specialized, fine-tuned NLLB-1.3B model. The proprietary GPT-5 model sets a high bar, establishing the state-of-the-art for single-call systems, achieving a COMET score of 73.0 vs. the 70.1 of the baseline.

Our two-stage RAG pipeline, however, decisively improves performance across all open-source models. The augmentation provides a substantial boost, with Llama-3-70B gaining over 5.5 points in both BLEU and COMET, while the smaller Qwen3-14B sees a similar increment in both metrics, standing as a highly competitive alternative to the much larger and slower 70B model (2,3X slowest). This confirms the efficacy of providing in-context examples to guide the refinement process. Crucially, our Llama-3-70B+RAG system emerges as the top-performing model overall, surpassing both the specialized NLLB baseline and the powerful GPT-5 on all semantic metrics and BLEU.

It is noteworthy that while RAG achieves the highest scores, the BLEU values for all systems remain in the mid-20s. This likely represents a ceiling imposed by the 19th-century English references. In contrast, the semantic metrics tell a more revealing story: high BERTScore ($>$0.92) and COMET ($>$0.72) values, indicating that the core meaning is being translated accurately, even if the surface form differs.

\subsection{Results on the Out-of-Domain Test Set}

\begin{table}[H]
\centering
\small
\begin{tabular}{lccccc}
\toprule
\textbf{System (on Chartres OOD Test Set)} & \textbf{BLEU}$\uparrow$ & \textbf{chrF++}$\uparrow$ & \textbf{METEOR}$\uparrow$ & \textbf{BERTScore}$\uparrow$ & \textbf{COMET}$\uparrow$ \\
\midrule
\textit{Baselines} \footnotesize{(GPU)} & & & & & \\
NLLB-1.3B (fine-tuned) & 24.04 & 48.59 & 0.546 & 0.918 & 0.697 \\
Llama-3.3-70B (zero-shot) & 28.17 & 52.68 & 0.587 & 0.927 & 0.714 \\
Qwen3-14B (zero-shot) & 26.78 & 50.95 & 0.571 & 0.924 & 0.710 \\
\midrule
\textit{Proprietary Models} \footnotesize{(API)} & & & & & \\
GPT-5 & 34.05 & 57.05 & 0.641 & \textbf{0.938} & \textbf{0.760} \\
\midrule
\textit{Our Open RAG Systems} \footnotesize{(GPU)} & & & & & \\
Llama-3.3-70B + RAG (temp 0.5) & \textbf{34.68} & \textbf{57.33} & \textbf{0.646} & 0.936 & 0.757 \\
Llama-3.3-70B + RAG (temp 0.0) & 34.28 & 57.00 & 0.642 & 0.934 & 0.753 \\
Qwen3-14B + RAG (temp 0.0) & 32.81 & 55.28 & 0.629 & 0.931 & 0.746\\
Qwen3-14B + RAG (temp 0.5) & 32.66 & 55.33 & 0.625 & 0.930 & 0.744 \\
\midrule
\textit{Single augmented System} \footnotesize{(GPU)} & & & & & \\
Llama-3.3-70B + $k=1$ (temp 0.0) & 30.28 & 54.60 & 0.612 & 0.930 & 0.742 \\
Qwen3-14B + $k=1$ (temp 0.0) & 29.15 & 52.63 & 0.605 & 0.928 & 0.735 \\

\bottomrule

\end{tabular}
\caption{Results on the out-of-domain Yves de Chartres test set. Is retrieval necessary? Comparing raw LLM → k=1 → full RAG shows that k=1 delivers the bulk of the improvement in semantics (COMET; BERT), while adding neighbors yields diminishing but positive returns in surface metrics (BLEU, chrF).}

\label{tab:chartres-results}
\end{table}

The OOD results (Table~\ref{tab:chartres-results}) crystallize the core findings of this paper and allow us to dissect the contributions of our pipeline's components. The first notable trend is the higher scores across the board, likely due to the contemporary English references that better match the LLMs' native style. Here, the zero-shot LLMs already outperform the fine-tuned NLLB model.

Mirroring the trend observed in in-domain, the refinement component provides incremental on top of draft-conditioning, but with a key insight: Simply providing the zero-shot Llama-3.3-70B model with the NLLB draft (the $k=1$ condition) boosts its COMET score from 0.714 to 0.742 (+2.8). This single draft is responsible for the vast majority of the semantic improvement, confirming that the specialized NMT model provides a powerful structural and semantic anchor.

The retrieval component ($k>1$) acts as a final polishing step. Adding the five retrieved examples lifts the COMET score further to a peak of 0.753, adding another 1 point. Notably, this RAG step provides an even larger boost to lexical metrics like BLEU (+4 points) and METEOR (+3.5 points). This is consistent with our hypothesis: the draft sets the semantic foundation, while the retrieved examples help the refiner polish lexical choice and improve semantic fidelity. Ultimately, our full pipeline achieves performance statistically on par with the GPT-5 benchmark. 

Furthermore, the results highlight the value of smaller models within this framework. While the Qwen3-14B+RAG system does not surpass the GPT-5 baseline, it remains highly competitive, achieving a COMET score of 0.746. This shows that the RAG methodology is not solely dependent on massive model scale, effectively elevating more accessible models into a competitive performance tier.

\subsection{Ablation: Fine-Tuning vs. Zero-Shot RAG for Generalization}
For contrast with zero-shot RAG, we conducted an ablation study. We fine-tuned the Llama and Qwen models on a small portion (10\%) of our parallel corpus and evaluated them on both test sets. As shown in Table~\ref{tab:finetuning-ablation}: while fine-tuning provides a modest BLEU/COMET in-domain boost, it yields negligible COMET change, a signature of overfitting on archaic style and vocabulary from the training data. By losing its broad generalization capabilities, its ability to adapt to the new domain and style of the Chartres letters is consequently harmed.

This finding supports our draft-based methodology. Instead of forcing the model to fine-tune on a specific style, our zero-shot RAG approach preserves the LLM's vast prior knowledge and guides it at inference time. This makes it a far more robust solution for real-world scenarios across diverse historical periods and textual genres.

\begin{table}[h!]
\centering
\small
\begin{tabular}{llcc@{\quad}cc}
\toprule
\multirow{2}{*}{\textbf{Model}} & \multirow{2}{*}{\textbf{Condition}} & \multicolumn{2}{c}{\textbf{Rosenthal (In-Domain)}} & \multicolumn{2}{c}{\textbf{Chartres (OOD)}} \\
\cmidrule(lr){3-4} \cmidrule(lr){5-6}
& & \textbf{BLEU}$\uparrow$ & \textbf{COMET}$\uparrow$ & \textbf{BLEU}$\uparrow$ & \textbf{COMET}$\uparrow$ \\
\midrule
\multirow{2}{*}{Llama3.3-70B} & Zero-shot & 20.50 & 0.688 & 28.17 & 0.712 \\
& FT (10\% corpus) & \textbf{22.15}\hspace{02pt}\rlap{\scriptsize\textcolor{darkgreen}{(+1.7)}} & \textbf{0.698}\hspace{2pt}\rlap{\scriptsize\textcolor{darkgreen}{(+1.0)}} & 25.28\hspace{2pt}\rlap{\scriptsize\textcolor{darkred}{(-2.9)}} & 0.712\hspace{2pt}\rlap{\scriptsize\textcolor{darkred}{(-0.01)}} \\
\midrule
\multirow{2}{*}{Qwen3-14B} & Zero-shot & 19.35 & \textbf{0.673} & \textbf{26.78} & \textbf{0.710} \\
& FT (10\% corpus) & \textbf{20.67}\hspace{2pt}\rlap{\scriptsize\textcolor{darkgreen}{(+1.3)}} & 0.688\hspace{2pt}\rlap{\scriptsize\textcolor{darkgreen}{(+1.5)}} & 23.92\hspace{2pt}\rlap{\scriptsize\textcolor{darkred}{(-2.9)}} &  0.708\hspace{2pt}\rlap{\scriptsize\textcolor{darkred}{(-0.02)}} \\
\bottomrule
\end{tabular}
\caption{Effect of fine-tuning on a small data portion. Scores on the in-domain test set are improved but generalization to the OOD set are unchanged, a classic sign of overfitting.}
\label{tab:finetuning-ablation}
\end{table}

\


\section{Qualitative Analysis}
\label{sec:qualitative_analysis}

To illustrate the qualitative impact of our pipeline, this section analyzes two challenging translations from our OOD dataset (Tables \ref{tab:qualitative_examples_a} and \ref{tab:qualitative_examples_b}). The examples were chosen to demonstrate how our method addresses distinct linguistic phenomena in Latin that cause predictable errors in base models. We compare the outputs of Llama3.3 and Qwen3 against a state-of-the-art proprietary benchmark (GPT-5).

\begin{table*}[h!]
\small
\centering
\begin{tabularx}{\textwidth}{@{} l >{\raggedright\arraybackslash}X @{}}
\toprule
\multicolumn{2}{@{}l}{\textbf{Example 1: Complex Pronoun Linkage}} \\
\midrule
\textbf{Source (Latin)} & \emph{inspirante Dei gratia et prosequente, decrevi pati pro lege Dei mei nec ulla ratione cogente volo ei esse consentaneus in culpa, qui nolo esse consors in poena.} \\
\textbf{Reference} & The grace of God inspiring and sustaining me, I have resolved to suffer for the law of my God, and no reason will compel me to agree in sin with one whose punishment I do not wish to share. \\ 
\midrule
\textbf{System} & \textbf{Prediction Output} \\
\midrule
Llama3.3-70B & Inspired by the grace of God and guided by it, I have decided to suffer for the law of my God and I do not wish to be in agreement with him in guilt for any reason, \textbf{I who do not wish} to be a partner in punishment. \\
\makecell[l]{Llama3.3-70B \\ + RAG} & By the inspiration and guidance of God's grace, I have resolved to suffer for the law of my God, and without being compelled by any reason, I am unwilling to be a partner in guilt \textbf{with whom I do not wish to} share the punishment. \\
\midrule
Qwen3-14B & Inspired by the grace of God and pursuing it, I have decided to suffer for the law of my God, and I do not wish to be in agreement with anyone in guilt, \textbf{who I do not wish} to be a companion in punishment. \\
\makecell[l]{Qwen3-14B \\ + RAG} & Inspired by the grace of God and following it, I have determined to suffer for the law of my God, and not being compelled by any reason, \textbf{I do not wish to be in agreement in sin with who} I do not wish to share the punishment. \\
\midrule
\makecell[l]{GPT-5 \\ (Benchmark)} & With the grace of God inspiring and accompanying me, I have resolved to suffer for the law of my God; nor, when no reason compels me, do I wish to be in league with him in guilt—\textbf{whose punishment I do not wish to share}. \\
\bottomrule
\end{tabularx}
\caption{Comparison of model outputs with and without our RAG augmentation. For each example, we highlight key phrases where RAG induces significant changes in syntactic or grammatical fidelity.}
\label{tab:qualitative_examples_a}
\end{table*}

\paragraph{Example 1: Complex Pronoun Linkage.}
The source in Table~\ref{tab:qualitative_examples_a} bundles two ablative absolutes (\emph{inspirante … prosequente; nec ulla ratione cogente}), a dative-anchored predicate (\emph{consentaneus + ei}), and a relative that must corefer to that dative (\emph{ei … qui}) within a negated volitional frame. Baselines struggle with this density: Llama3.3-70B (base) gets the gist but simplifies the coercion template (“for any reason”) and produces a garden-path relative (“I who do not wish”). Qwen3-14B (base) literalizes \emph{prosequente} (“pursuing it”) and repeats the same mis-scoped relative.

With augmentations, Llama3.3-70B correctly restores the coercion template and the “accompanying/guidance” sense of \emph{prosequente}; crucially, it resolves the pronominal linkage with a correct syntactic form (“with whom I do not wish to share the punishment”) avoiding earlier semantic drift. By contrast, Qwen3-14B + RAG learns the coercion template, but tangles the relative (“in agreement in sin with who I do not wish to share…”), showing a who/whom error and a redundant preposition.

GPT-5 serves as a fluency benchmark, resolving the linkage with an idiomatic possessive (“…whose punishment…”). Our augmented LLama3.3 output produces an equally correct but more literal rendering (“…with whom…”) closer to the source's syntactic structure, a quality often preferred for scholarly use.

\begin{table*}[h!]
\small
\centering
\begin{tabularx}{\textwidth}{@{} l >{\raggedright\arraybackslash}X @{}}
\toprule
\multicolumn{2}{@{}l}{\textbf{Example 2: Passive Voice Fidelity}} \\
\midrule
\textbf{Source (Latin)} & \emph{Ita dicens, Minervam quidem non doceo, a qua magis doceri indigeo, sed fraternis profectibus congaudens, bonis principiis meliores exitus apponi desidero.} \\
\textbf{Reference} & In speaking thus, I am assuredly not teaching Minerva, by whom I rather need to be taught, but in rejoicing in the progress of the brethren, I desire for better ends to be joined to good beginnings. \\
\midrule
\textbf{System} & \textbf{Prediction Output} \\
\midrule
Llama3.3-70B & Saying this, I do not teach Minerva, from whom I more \textbf{need to learn}, but rejoicing in fraternal progress, I desire better outcomes to be joined to good principles. \\
\makecell[l]{Llama3.3-70B \\ + RAG} & Thus saying, I do not teach Minerva, from whom I have greater \textbf{need to be instructed}, but rejoicing in the progress of my brothers, I desire that better outcomes be assigned to good beginnings. \\
\midrule
Qwen3-14B & So saying, I do not teach Minerva, from whom I \textbf{need to learn more}, but being delighted with the good progress of my brother, I desire that better results may follow from good beginnings. \\
\makecell[l]{Qwen3-14B \\ + RAG} & Thus speaking, I do not teach Minerva, from whom I have greater \textbf{need to be taught}; but rejoicing in my brothers' progress, I desire better results to be attached to good beginnings. \\
\midrule
\makecell[l]{GPT-5 \\ (Benchmark)} & In saying this I do not teach Minerva—by whom I rather \textbf{need to be taught}—but, rejoicing in a brother’s progress, I desire that better outcomes be added to good beginnings. \\
\bottomrule
\end{tabularx}
\caption{Comparison of model outputs with and without our RAG augmentation. For each example, we highlight key phrases where RAG induces significant changes in syntactic or grammatical fidelity.}
\label{tab:qualitative_examples_b}
\end{table*}

\paragraph{Example 2: Enforcing Grammatical Fidelity.} The second example in Table~\ref{tab:qualitative_examples_b}  tests a different cluster: fidelity to the passive voice (a humility topos with \emph{magis doceri indigeo}), a dative governed by \emph{congaudere} (\emph{fraternis profectibus}), and an additive construction with \emph{apponi} + dative (\emph{meliores exitus … bonis principiis}), while also requiring the correct resolution of the polysemy in \emph{principiis} (“beginnings,” not the doctrinal “principles”).

The base models stumble in predictable simplifications: both Llama3.3 and Qwen3 switch the passive infinitive to active (“need to learn”), Llama3.3 reads \emph{principiis} as “principles”, and Qwen3 turns the additive relation into a causal one (“follow from”).

The augmentations correct these deficits. Both Llama3.3 and Qwen3 restore the passive (“need to be taught”), keep the dative roles intact, and map \emph{apponi} to an additive verb targeting a dative goal (“assigned / attached to good beginnings”). GPT-5 is also correct on passive and additivity; its singular “\emph{a brother’s}” is a harmless stylistic choice relative to the Latin plural.

Where the first example showed our pipeline resolving a critical dependency failure, this one demonstrates its ability to provide the lexical and syntactic scaffolding needed to correct more pervasive patterns of simplification. With this guidance, both open-source models reach grammatical parity with the benchmark on all key features. The remaining differences in verb choice (assigned / attached / added) are again stylistic rather than philological. 

\textbf{Two Paths to Correctness.} While quantitative metrics show benchmark parity, the paths to correctness differ. Across both examples, retrieval consistently stabilizes the parts of Latin that most often fail in open models: clause scope, case governance, and formulaic patterns, bringing COMET scores in line with closed models (76.0 vs. 75.7 in OOD). Proprietary models often retain a lead in idiomaticity; while RAG-augmented outputs tend to preserve the source’s syntactic structure. In other words, our pipeline does not replace strong closed systems; it complements them by offering a source-faithful alternative at comparable adequacy, thereby adding stylistic diversity without sacrificing correctness. 

\section{Discussion}

Our results demonstrate that the draft-based refinement pipeline, enhanced by retrieval augmentation (RAG) can elevate open-source models to a level of performance that is quantitatively on par with, and qualitatively distinct from, state-of-the-art proprietary systems. This section discusses the broader implications of these findings.

\textbf{The Two Paths to Correctness.}
The central insight from our analysis is that benchmark parity does not imply stylistic identity. The qualitative comparison reveals that our retrieval augmented models and GPT-5 follow "two paths to correctness." GPT-5 excels at producing dynamic, idiomatic translations that prioritize fluency for a modern reader. Our pipeline, by contrast, consistently yields translations that are more structurally faithful to the Latin source. One might interpret the greater structural fidelity of our augmented outputs as a sign of rigidity compared to GPT-5's dynamic fluency. However, we argue this is a feature, not a limitation. For scholarly contexts or legal applications, this "rigidity" translates to traceability and philological precision, demonstrating that RAG can be used to instill specific, desirable constraints on a model's output.

\textbf{Architectural Synergy and the Role of the Drafter.}
The success of our pipeline is rooted in its architectural synergy. The fine-tuned NLLB model acts as a specialized "navigator," producing syntactically plausible drafts that anchor the much larger LLM refiners in the correct semantic space. The LLM then acts as a general reasoner, applying its vast pre-trained knowledge and linguistic fluency to polish this draft. Our methodological choice to use NLLB drafts as RAG examples helps the refiner to be an expert post-editor, specializing it on-the-fly to the specific error patterns and style of the drafter.

\textbf{The NLLB Draft as an "Ad Verbum" Anchor.}
A deeper reason for our pipeline's success lies in the NLLB draft's function as an \emph{ad verbum} (literal) anchor. The fine-tuned NMT model produces structurally conservative drafts that, while less fluent than a massive LLM, faithfully preserve core Latin syntax (case roles, clause dependencies). This aligns with the reasoning of Rosu (2025) \cite{rosu-2025-litera}: a literal first pass constrains the LLM's interpretative role. The draft provides the syntactic core, while examples retrieved from our vast diachronic corpus help when surface similarity or terminological consistency matters. This theory is empirically supported by our OOD results, where the NLLB draft alone accounted for the majority of the semantic performance gain over the zero-shot baseline.

\textbf{Limitations and Future Work.}
This study has several limitations that open avenues for future research. Our pipeline's performance is contingent on the quality of the retrieved examples; future work could explore more robust retrieval, filtering mechanisms, and new "thinking" capacities of LLMs. While we demonstrate strong performance on Latin-to-English, testing this architecture on other low-resource or morphologically-rich languages is a logical next step. Furthermore, the emergence of two distinct, high-quality translation styles suggests a promising direction for controllable MT, where a user could specify their desired level of "idiomaticity" vs. "fidelity" at inference time. Finally, while our ablation confirms the value of the draft-based approach over fine-tuning, a more granular analysis, such as a k-sweep to measure the marginal contribution of each retrieved neighbor, could further clarify the trade-offs between performance and latency.

\section{Conclusion}

We addressed the challenge of high-fidelity Latin translation by developing a reproducible, draft-based refinement pipeline, enhanced with retrieval augmentation. Our analysis reveals that a high-quality initial draft provides the primary boost in semantic accuracy, while the retrieval of in-context examples acts as a final polish agent for lexical choice and fluency, together elevating open-source LLMs to quantitative parity with a top-tier proprietary benchmark. More significantly, our qualitative analysis demonstrates that our system produces a philologically grounded alternative to the more idiomatic output of proprietary models. This work validates a viable path toward creating diverse, specialized, and state-of-the-art translation tools outside of closed ecosystems, offering researchers a meaningful choice in how they engage with texts about the past.


\section*{Ethics Statement}
Our system is designed for translating historical text. We acknowledge that historical documents, including Latin texts, may contain outdated or offensive viewpoints (e.g., colonial or sexist language in Victorian translations, or propagandistic statements in ancient Roman texts). Our approach focuses on literal translation and does not aim to filter or alter such content. Users of the system for public-facing translations should be aware of the potential for problematic content and might need to add post-processing or content warnings. Finally, by using open-source models and data, we adhere to the ethos of transparency and reproducibility, and avoid ethical concerns associated with proprietary AI deployment (such as hidden biases introduced by unseen training data). 

\section*{Reproducibility Statement}
We have described the data sources and training procedures in detail. In Section 3, we include hyperparameter settings, algorithms, and examples of prompts used. The release of models and the LoRA adapter weights for the models can be found on our Huggingface repository \href{https://huggingface.co/magistermilitum}{https://huggingface.co/magistermilitum}. The code to deploy our RAG system can be forked from our Gitlab: \href{https://gitlab.com/magistermilitum}{https://gitlab.com/magistermilitum}

\appendix

\bibliographystyle{acm}
\bibliography{references}

\begin{thebibliography}{10}

\bibitem{bge_m3}
{\sc Chen, J., Xiao, S., Zhang, P., Luo, K., Lian, D., and Liu, Z.}
\newblock Bge m3-embedding: Multi-lingual, multi-functionality, multi-granularity text embeddings through self-knowledge distillation, 2023.

\bibitem{opusbible}
{\sc Christodouloupoulos, C., and Steedman, M.}
\newblock A massively parallel corpus: the bible in 100 languages.
\newblock {\em Language resources and evaluation 49}, 2 (2015), 375--395.

\bibitem{nlbb2022}
{\sc Costa-Juss{\`a}, M.~R., Cross, J., {\c{C}}elebi, O., Elbayad, M., Heafield, K., Heffernan, K., Kalbassi, E., Lam, J., Licht, D., Maillard, J., et~al.}
\newblock No language left behind: Scaling human-centered machine translation.
\newblock {\em arXiv preprint arXiv:2207.04672\/} (2022).

\bibitem{dubey2024llama}
{\sc Dubey, A., Jauhri, A., Pandey, A., Kadian, et~al.}
\newblock The llama 3 herd of models.
\newblock {\em arXiv e-prints\/} (2024), arXiv--2407.

\bibitem{fischer2022machine}
{\sc Fischer, L., Scheurer, P., Schwitter, R., and Volk, M.}
\newblock Machine translation of 16th century letters from latin to german.
\newblock In {\em Proceedings of the Second Workshop on Language Technologies for Historical and Ancient Languages\/} (2022), pp.~43--50.

\bibitem{Yves_Chartres}
{\sc Giordanengo, G.}
\newblock Lettres d'{Yves de Chartres}.
\newblock \url{https://telma.irht.cnrs.fr/chartes/yves-de-chartres/}, jun, 2017.
\newblock TELMA.

\bibitem{iyer2024quality}
{\sc Iyer, V., Malik, B., and Stepachev, e.~a.}
\newblock Quality or quantity? on data scale and diversity in adapting large language models for low-resource translation.
\newblock In {\em Proceedings of the Ninth Conference on Machine Translation\/} (2024), pp.~1393--1409.

\bibitem{kocmi2021one}
{\sc Kocmi, T., and Bojar, O.}
\newblock One model to learn them all: Multilingual transfer with sparse mixture of experts.
\newblock In {\em EMNLP\/} (2021).

\bibitem{gpt-5}
{\sc OpenAI}.
\newblock {GPT-5} system card.
\newblock \url{https://cdn.openai.com/gpt-5-system-card.pdf}, august, 2025.

\bibitem{papineni2002bleu}
{\sc Papineni, K., Roukos, S., Ward, T., and Zhu, W.-J.}
\newblock {BLEU}: a method for automatic evaluation of machine translation.
\newblock In {\em ACL\/} (2002).

\bibitem{popovic2015chrf}
{\sc Popovi{\'c}, M.}
\newblock chr{F}: character n-gram f-score for automatic mt evaluation.
\newblock In {\em Proceedings of the tenth workshop on statistical machine translation\/} (2015), pp.~392--395.

\bibitem{post2018call}
{\sc Post, M.}
\newblock A call for clarity in reporting bleu scores.
\newblock In {\em WMT\/} (2018).

\bibitem{rei2022comet}
{\sc Rei, R., De~Souza, J.~G., Alves, D., Zerva, C., Farinha, A.~C., Glushkova, T., Lavie, A., Coheur, L., and Martins, A.~F.}
\newblock Comet-22: Unbabel-ist 2022 submission for the metrics shared task.
\newblock In {\em Proceedings of the Seventh Conference on Machine Translation (WMT)\/} (2022), pp.~578--585.

\bibitem{rosenthal2023latin}
{\sc Rosenthal, G.}
\newblock {\em Machina cognoscens: Neural machine translation for latin, a case-marked free-order language}.
\newblock PhD thesis, Master’s thesis, University of Chicago, 2023.

\bibitem{rosu-2025-litera}
{\sc Rosu, P.}
\newblock {LITERA}: An {LLM} based approach to {L}atin-to-{E}nglish translation.
\newblock In {\em Findings of the Association for Computational Linguistics: NAACL 2025\/} (Albuquerque, New Mexico, Apr. 2025), L.~Chiruzzo, A.~Ritter, and L.~Wang, Eds., Association for Computational Linguistics, pp.~7781--7794.

\bibitem{volk-etal-2024-llm}
{\sc Volk, M., Fischer, D.~P., Fischer, L., Scheurer, P., and Str{\"o}bel, P.~B.}
\newblock {LLM}-based machine translation and summarization for {L}atin.
\newblock In {\em Proceedings of the Third Workshop on Language Technologies for Historical and Ancient Languages (LT4HALA) @ LREC-COLING-2024\/} (Torino, Italia, May 2024), ELRA and ICCL, pp.~122--128.

\bibitem{white2015blitz}
{\sc White, J.~F.}
\newblock Blitz latin revisited.
\newblock {\em Journal of Classics Teaching 16}, 32 (2015), 43--49.

\bibitem{yang2025qwen3}
{\sc Yang, A., Li, A., Yang, B., Zhang, B., Hui, B., Zheng, B., Yu, B., Gao, C., Huang, C., Lv, C., et~al.}
\newblock Qwen3 technical report.
\newblock {\em arXiv preprint arXiv:2505.09388\/} (2025).

\bibitem{zhangbertscore}
{\sc Zhang, T., Kishore, V., Wu, F., Weinberger, K.~Q., and Artzi, Y.}
\newblock {BERTS}core: Evaluating text generation with bert.
\newblock In {\em International Conference on Learning Representations\/} (2020).

\end{thebibliography}

\newpage

\end{document}